# A Novel Cluster Detection of COVID-19 Patients and Medical Disease Conditions Using Improved Evolutionary Clustering Algorithm Star

Bryar A. Hassan[1,2*], Tarik A. Rashid[3], Hozan K. Hamarashid[4]

[1]Department of Computer Networks, Technical College of Informatics, Sulaimani Polytechnic University, Sulaimani 46001, Iraq
[2]Kurdistan Institution for Strategic Studies and Scientific Research, Sulaimani 46001, Iraq
Email (Corresponding)*: bryar.hassan@kissr.edu.krd

[3]Computer Science and Engineering Department, University of Kurdistan Hewler, Iraq
Email: tarik.ahmed@ukh.edu.krd

[4]Information Technology Department, Computer Science Institute, Sulaimani Polytechnic University, Sulaimani 46001, Iraq
Email: hozan.khalid@spu.edu.iq

## Abstract

With the increasing number of samples, the manual clustering of COVID-19 and medical disease data samples becomes time-consuming and requires highly skilled labour. Recently, several algorithms have been used for clustering medical datasets deterministically; however, these definitions have not been effective in grouping and analysing medical diseases. The use of evolutionary clustering algorithms may help to effectively cluster these diseases. On this presumption, we improved the current evolutionary clustering algorithm star (ECA*), called iECA*, in three manners: (i) utilising the elbow method to find the correct number of clusters; (ii) cleaning and processing data as part of iECA* to apply it to multivariate and domain-theory datasets; (iii) using iECA* for real-world applications in clustering COVID-19 and medical disease datasets. Experiments were conducted to examine the performance of iECA* against state-of-the-art algorithms using performance and validation measures (validation measures, statistical benchmarking, and performance ranking framework). The results demonstrate three primary findings. First, iECA* was more effective than other algorithms in grouping the chosen medical disease datasets according to the cluster validation criteria. Second, iECA* exhibited the lower execution time and memory consumption for clustering all the datasets, compared to the current clustering methods analysed. Third, an operational framework was proposed to rate the effectiveness of iECA* against other algorithms in the datasets analysed, and the results indicated that iECA* exhibited the best performance in clustering all medical datasets. Further research is required on real-world multi-dimensional data containing complex knowledge fields for experimental verification of iECA* compared to evolutionary algorithms.

# 1. Introduction

Data mining techniques have a crucial role in decision-making and prediction. In particular, clustering organises observations in a dataset by grouping related observations in the same cluster and dissimilar observations in distinct clusters. Clustering algorithms are used in several areas, including medical patient records, web text mining, and business market analysis. Numerous clustering algorithms have been suggested, but each technique is mainly devoted to a particular form of a problem [1]. For example, [2] concluded that K-means exhibits reduced performance on datasets with a large number of clusters, small cluster sizes, or cluster imbalances. Clustering algorithms have different performances in different datasets and real-world applications. Therefore, it is essential to analyse the sensitivity of an algorithm to a range of benchmarking and real-world problems. Thus, almost all clustering techniques have a range of disadvantages. First, it is difficult to determine the optimal number of clusters [3]. Second, clustering algorithms are susceptible to the random sorting of cluster centroids. Selecting insufficient cluster centroids can easily result in ineffective clustering solutions [4]. Third, because virtually any clustering algorithm involves a hill-climbing process to achieve its goal, local optima will easily be stuck, resulting in suboptimal clustering results [5]. Fourth, the evidence cannot be isolated from noise and outliers; we conclude that clusters have common distributions and near-identical masses. As a result, noise and outliers lead to excellent clustering outcomes, where noise sources and outliers are present. Fifth, there are few studies demonstrating the vulnerability of clustering algorithms to dataset cohorts and real-world implementations. Finally, most of the previous algorithms use a deterministic-based approach [1]; thus, their clustering results are primarily dependent on their initial states and inputs, and the output generation process is affected by the starting conditions and initialisation parameters. In addition, clustering algorithms are incapable of quickly capturing both local and global optimal spaces [6].

Furthermore, clustering is important in helping medical experts group a specific type of disease. Current clustering algorithms have been developed for various real-world applications, such as science, image processing, medicine, and decision-making agents [7,8]. For instance, dataset samples provided via diagnosis in the medical field are required for disease analysis, and are analysed by a doctor or pharmacist to determine the stage of the disease. As the number of patients increases, more time is required to examine the samples. Hence, a systematic method is needed to automatically or semi-automatically evaluate the sample dataset for each patient. The data samples of medical conditions can be categorised by applying a systematic method involving clustering algorithms.

There has been insufficient research to effectively cluster COVID-19 and other medical disease datasets using clustering algorithms. As an exception, the study in [9] introduced an overlapping k-means algorithm for medical applications, despite the limitations of this algorithm. Another study [10] examined the potential of extending ensemble clustering methods to the field of medical diagnostics.

Recently, a new evolutionary clustering algorithm, called ECA*, was proposed in [7] for heterogeneous and numerical datasets. ECA* was developed based on statistical and evolutionary algorithms [6,11]. This newly introduced algorithm was examined using state-of-the-art clustering algorithms, and the results indicated that better clustering results were obtained with ECA* compared to other competitive techniques. Moreover, an adaptive version of ECA* has been developed to reduce the size of concept hierarchies from corpora [12]. According to the same study, the resulting lattice was homoeomorphic to the original one, preserving the structural relationship between the two definition lattices through the prism of the experiment and outcome analysis. Compared to the basic concept lattice, this resemblance between the two lattices maintained the consistencies of the resulting definition hierarchies by 89%, with a loss of 11%. Thus, the quality of the resulting concept hierarchies is promising. Nonetheless, ECA* do not have limitations. Adapting ECA* to different practical problems is a challenge that needs to be addressed. Furthermore, ECA* can be exploited under the assumption of no prior input information. As a result, we used an enhanced ECA* to create a proper exercise protocol and a novel research method for blended multi-variance datasets. Thus, we improved ECA* to effectively cluster real datasets in COVID-19 and other conditions. The proposed algorithm is called improved ECA* (iECA*). Our newly introduced algorithm has four significant advantages over the standard ECA*: (i) the elbow technique is used to determine the optimum number of clusters; (ii) the input datasets are cleaned and processed as part of the iECA*; (iii) unlike ECA*, iECA* works on multivariate and domain-theory datasets with different attribute characteristics, such as integer, real, and categorical data attributes; (iv) iECA* can be used in real-world applications for diagnosing medical disease datasets.

This study aims to effectively cluster real-world datasets involving COVID-19 symptom checkers, liver disorders, diabetes, and kidney and heart diseases using iECA*. To evaluate the effectiveness of iECA* on the actual medical data, we examined iECA* against seven modern algorithms (ECA*, genetic algorithm for clustering++ (GENCLUST++), k-nearest neighbours algorithm (KNN), deep KNN, learning vector quantisation (LVQ), support vector machine (SVM), and artificial neural network (ANN)). In addition, the performance of the methods was compared and analysed.

The remainder of this paper is organised as follows: Section 2 reviews previous works related to evolutionary clustering algorithms; Section 3 presents the newly suggested algorithm based on ECA*; Section 4 discusses the research methods of the study; in Section 5, the results of the algorithms are deliberated based on performance and validation measures; finally, Section 6 draws the concluding remarks and describes future research efforts.

## 2. Related Works

Numerous clustering methods have been developed over the last few years. ECA* is one of the most recently developed evolutionary clustering algorithms that addresses the limitations of the currently available clustering methods. ECA* is an ensemble evolutionary clustering algorithm used for clustering heterogeneous and multi-featured real datasets and practical applications, and it integrates several approaches [7][13]: statistical, heuristic, and evolutionary methods. Thus, it has been used for analysing multi-featured and heterogeneous datasets. ECA* comprises five parts: (i) initialisation, (ii) clustering I, (iii) mut-over, (iv) clustering II, and (v) evaluation. Fig. 1 shows the detailed flowchart of ECA* for the numerical datasets. In addition to that, the pseudo-code of ECA* is available in [14].

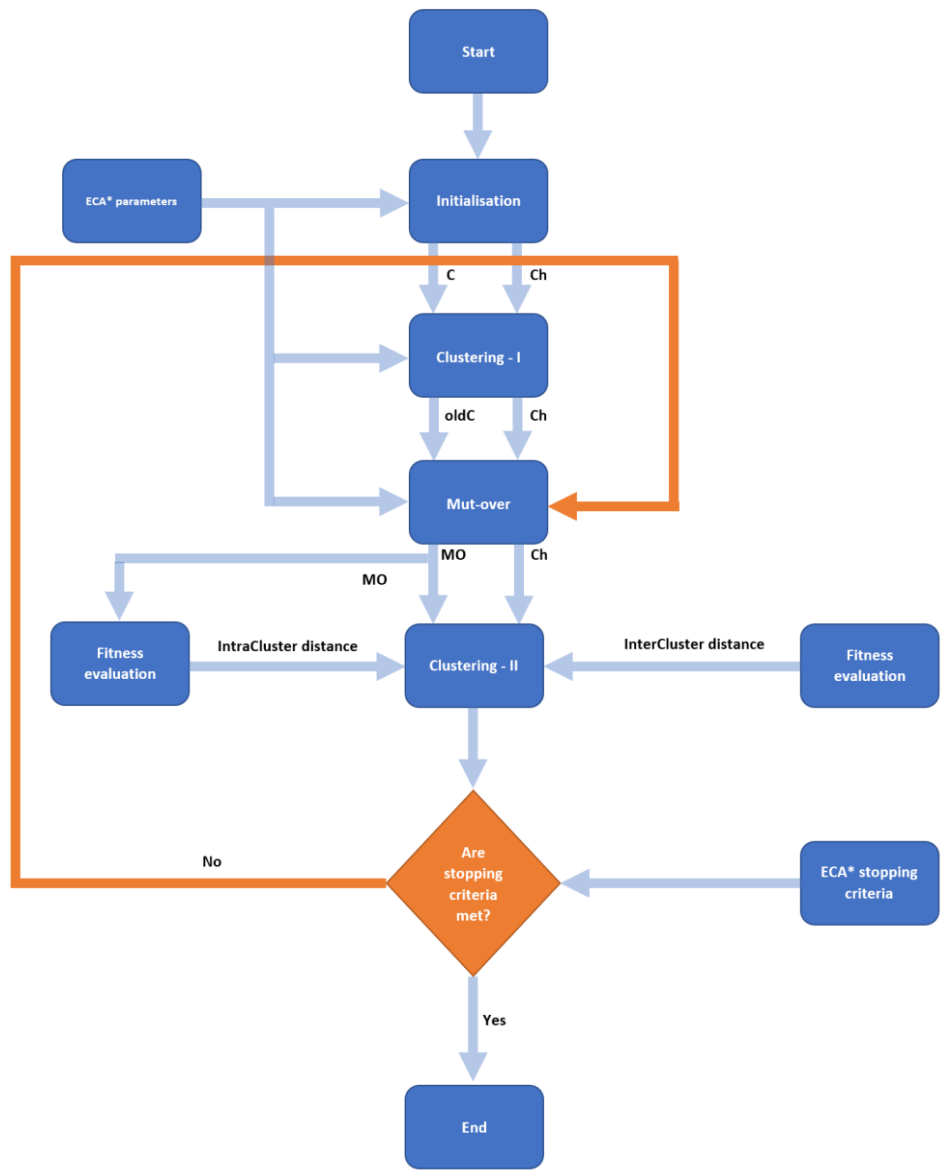

**Fig. 1. The detailed flowchart of ECA* for numerical datasets (adapted from [7])**

The components and mathematical formulations of ECA* are explained below:

**1. Initialisation.** The input dataset consists of many records, and each one has several numerical properties. Assume that the input data (dataset) is represented by N chromosomes $Ch_i$, and each chromosome contains a collection of genes ($G_{i0}$, $G_{i1}$, $G_{i2}$,...$G_{ij}$).

For i=0, 1, 2,....., N and j=0, 1, 2,....., D, where N and D are the number of records and attributes, respectively.

At this step, the following parameters should be initialised:

  A. Social class ranks (*S*) and number of clusters (K), as presented in Equation (1):
$$K = S^j \tag{1}$$
  B. Minimum cluster density threshold ($C^{dth}$).
  C. Random walk (*F*).
  D. Crossover type ($C^{type}$).

Subsequently, the percentile rank ($P_{ij}$) for each data point and its average percentile rank ($P_i$) for each chromosome should be determined. Finally, each chromosome is assigned to a cluster (*K*) based on its rank.

**2. Clustering-I.** This component consists of the following steps:

  A. The number of real clusters is computed using Equation (2):

$$K^{new} = K - K^{empty} - K^{dth} \tag{2}$$

  where $K^{empty}$ is the number of empty clusters and $K^{dth}$ is the number of low-density clusters.

  B. The initial cluster centroids are calculated using Equation (3):
$$C_i = mean\ quartile\ (Ch_{ij}), \tag{3}$$

  where $Ch_{ij}$ is a set of chromosomes.

  C. The old cluster centroids are computed using Equation (4):

$$oldC_i \sim U\ (lQuartile, uQuartile) \tag{4}$$

  D. For each cluster, the intraCluster and oldIntraCluster are determined. Similarly, for the current clustering solution, the interCluster and oldInterCluster are calculated. Equations (5) and (6) present the mathematical formulas of intraCluster and interCluster calculations for clusters A and B, respectively:

$$\text{intraclass of } A = \frac{1}{|A|.(|A|-1)} \sum_{\substack{x,y \in A \\ x \neq y}}^{n} \{d(x,y)\} \tag{5}$$

$$\text{InterCluster of } (A, B) = \frac{1}{|A|+|B|} \left\{ \sum_{x \in A} d(x, v_b) + \sum_{y \in B} d(y, v_a) \right\} \tag{6}$$

where $v_a = \frac{1}{|A|} \sum_{x \in A} x$ and $v_b = \frac{1}{|B|} \sum_{y \in B} y$.

E. Finally, new cluster centroids are calculated as presented in Equation (7):

$$newC_i := \begin{cases} C_i & intraCluster_i < oldIntraCluster_i \\ oldC_i & otherwise \end{cases} \tag{7}$$

**3. Mut-over.** This strategy consists of a recombination operator of mutation and crossover.

A. Mutation: Mutate each cluster centroid of *i* to relocate it to the densest region of the cluster. The cost of mutating each cluster is computed using Equation (8):

$$Mutant_i := C_i + F(HI) \tag{8}$$

where *F* is the random walk initialisation and *HI* is the historical information, which is calculated as expressed in Equation (9):

$$HI := \begin{cases} oldC_i - C_i & If\ intraCluster_i < oldIntraCluster_i \\ C_i - oldC_i & Otherwise \end{cases} \tag{9}$$

B. Crossover: Using a uniform crossover operator, a new cluster centroid is created from the current and previous cluster centroids. The new cluster centroid (newCi) for cluster *i* is computed as illustrated in Equation (10):

$$newC_i := oldC_i + C_i \tag{10}$$

In addition, there is a switch operator between crossover and mutation in ECA* that generates the final trial of centroids between mutation and crossover using objective functions, as defined in Equation (11):

$$MO_i := \begin{cases} Mutant_i\ if\ oldInterCluster_i > interCluster_i \\ newC_i & Otherwsie \end{cases} \tag{11}$$

As a result of the mutation technique, certain cluster centroids produced after the crossover process may exceed their search space constraints. The mut-over operator in ECA* is comparable to the boundary control technique in backtracking search optimisation algorithm [15]. Moreover, the boundary control method of ECA* successfully creates population diversity, which guarantees that effective searches for clustering and cluster centroid findings are generated.

**4. Clustering-II.** After generating the mut-over operator, the clusters are merged according to their diversity, and the cluster centroids are recalculated. At this stage, the clusters that are close are merged. In turn, the distance between closed clusters should be considered. Several distance metrics exist between clusters that are frequently used [16]. The minimum and maximum distance methods, centroid distance method, and cluster-average method are used in this algorithm [17]. Therefore, the minimum distance between two clusters can be defined in Equation (12) [17] as follows:

$$D_{min}(X,Y) := \min\{d(A_x, A_y)\} \quad (12)$$

where $A_x \in X, A_y \in Y$.

The diversity between Ci and Cj is represented in Equation (13) [17] as follows:

$$\sigma_{ij} = \min\{(D_{min}(C_i, C_j) - R(C_i))(D_{min}(C_i, C_j) - R(C_j))\} \quad (13)$$

where R(ci) and R(Cj) are the average distances of the intraCluster and $D_{min}(C_i, C_j)$ is the minimum distance between Ci and Cj.

The criteria for merging two clusters should be one of the following:

A. If the result ($\sigma \leq 0$) is less than or equal to zero, these two classes are closely related and highly interrelated. As a result, classes Ci and Cj may be combined into a single class (Cij). That is, once a low-density class is created, it will be empty;

B. The average intraCluster distance between these two clusters is smaller than the shortest distance. This principle implies that Ci and Cj continue to exist as distinct clusters.

Finally, the number of clusters is recalculated to eliminate empty clusters generated owing to the lower-density clusters.

**5. Fitness evaluation.** This component is used to determine the clusters. The interCluster and intraCluster distances of the produced clusters are used as inputs, and the output is the fitness. The algorithm will terminate if the interCluster distance achieves its most significant value and the intraCluster distance reaches its minimum value (optimal). However, this criterion is not feasible. If the following conditions are satisfied, the halting requirements are satisfied:

A. The method completes the number of iterations specified in the initialisation;
B. The interCluster and intraCluster values remain constant during each cycle. The value of the interCluster does not increase, and the value of the intraCluster does not decrease throughout each iteration.

# 3. iECA*

One of the advantages of ECA* is the use of stochastic and random procedures. The stochastic method* of ECA is advantageous because it strikes a balance between navigating the search space and using the search space learning process to focus on global and local optima. Moreover, the superior efficiency of ECA* is a product of the use of operators with meta-heuristic algorithms in three aspects [7,11]. First, the adaptive control parameter (F) is implemented using Levy flight optimisation to balance the exploitation and exploration of the algorithm. Second, to enhance the capacity of the cluster centroids for learning and determine the optimal cluster centroids, the cluster centroids learn information from historical cluster centroids (HI). Mut-over may also be used to describe a recombination technique involving mutation-crossover. Third, mut-over can resolve the issue of global and/or local optima that might arise in other clustering techniques [18] when F and HI are used. This recombined approach confers consistency and robustness to the proposed algorithm [19]. As a result, these methods maintain an adequate balance between global and local optima.

However, the experimental findings of [7] demonstrate some shortcomings of the ECA*:

1. Finding the ideal pre-defined value for the variables of ECA*, such as the number of social class levels and the cluster density criterion, is challenging. Selecting the ideal number of social class ranks may preclude determining the ideal number of clusters;

2. Changing the number of social class ranks may alter the definition of the cluster threshold density. A limited number of social class ranks may result in a small number of clusters and a high threshold for cluster density. In contrast, many social class ranks may result in a large number of clusters and a low threshold for cluster density. Because social class rankings and the cluster density criterion are pre-defined values, balancing these two factors might be complicated;

3. ECA* has been previously used for numerical data, but it has not been used for multi-variance data and real-world applications;

4. Data cleaning and processing is not considered a part of ECA*.

In this research, we improved the ECA* in four aspects:

1. We utilise the elbow method to find the ideal number of clusters. The elbow method is perhaps the most well-known approach for determining the optimum cluster number. This method is a heuristic method used in cluster analysis when calculating the number of clusters in each dataset [20]. The method relies on the number of clusters and involves plotting the explained variance and selecting the elbow of the curve to use the cluster numbers. The same approach can be used to select the number of parameters, such as the number of principal components used to define the data collection, in other data-driven models. As stated in [15], the elbow method has a faster execution time than other methods (gap statistic, silhouette coefficient, canopy) to find the optimal number of clusters;

2. The input dataset is cleaned and processed in two steps. (i) Data cleaning: The input dataset may include many unnecessary and missing elements. Data cleansing is performed to address this issue, and includes the management of missing data and noisy data. (ii) Dataset processing: This step is used to convert the data into a format suitable for mining. This process is accomplished using normalisation and de-normalisation processes to scale data values within a defined range, such as -1.0 to 1.0 or 0.0 to 1.0;

3. Unlike ECA*, iECA* is applied to multivariate and domain-theory real datasets with different attribute characteristics, such as integer, real, and categorical data attributes;

4. iECA* is used for real-world applications in clustering COVID-19 and medical disease datasets.

On this premise, iECA* includes five parts: (i) initialisation, (ii) pre-processing, (iii) realignment of mutation and crossover, (iv) post-processing, and (v) evaluation. Algorithm 1 shows the pseudo-code of the iECA* for COVID-19 and medical disease records.

**Algorithm 1. Pseudo-code of iECA***

```
Input: S, K, C^dth, F, C^type, maxcycle, P, N, D, Ch_ij
Output: A set of classes (MO), class centroids (C)
 1  //1. INITIALISATION;
 2  K = S^D for i ← 0 to N do
 3      for j ← 0 to D do
 4          Ch_ij = Dataset_ij
 5      end
 6  end
 7  for i ← 0 to N do
 8      for j ← 0 to D do
 9          P_ij = Percentilisation (Ch_ij)
10      end
11  end
12  for iteration ← 1 to maxcycle do
13      //2. PRE-PROCESSING;
14      K^new = K - K^empty ;
15      for i ← 0 to N do
16          if ((Ch_i belongs to P_i ) then
17              if ((Ch_i is not missing and/or noisy data ) then
18                  Normalise C_i; ;
19                  C_i := MeanQuartilei(Ch_i);
20              end
21          end
22      end
23      // 3. MUT-OVER;
24      for i ← 0 to N do
25          if ((intraClass_i ¡ oldIntraClass_i ) then
26              HI_i := oldC_i - C_i;
27          end
28          else
29              HI_i := C_i - oldC_i;
30          end
31          Mutant = C_i + F (H_i);
32          newC_i = uniformCrossover (oldC_i + C_i);
33      end
34      Call Mut-over strategy;
35      // 4. POST-PROCESSING;
36      for i ← 0 to K do
37          for j ← 0 to K do
38          end
39          De-normalise C_i; ;
40      end
41      // 5. FITNESS EVALUATION;                  1
42      if (oldInterClass_i = interClass_i) AND (oldIntraClass_i = intraClass_i) then
43          // Export the cluster centroids and their observations;
44      end
45  end
```

The pseudo-code mut-over strategy for iECA* is also presented in Algorithm 2.

**Algorithm 2. Pseudo-code of the mut-over strategy of iECA***

```
Input: K, MO, Mutant, C, InterClass, oldInterClass
Output: MO
1 for i ← 0 to K do
2     if (oldInterClass_i > interClass_i) then
3         MO_i := Mutant_i;
4     end
5     else
6         MO_i := C_i;
7     end
8 end
```

Furthermore, the iECA* Java code is available in [21] and [22]. Also, the components and mathematical formulations of iECA* are presented below.

**1. Initialisation.** This component, a categorised or/and numerical input dataset (N × D), is initialised into the algorithm. Subsequently, the parameters listed below should be initialised.

   A. Number of clusters (K).

   B. Minimum cluster density threshold ($C^{dth}$).

   C. Random walk (F).

   D. Crossover type ($C^{type}$).

Subsequently, the percentile rank ($P_{ij}$) for each data point and its average percentile rank ($P_i$) for each chromosome should be determined. Finally, each chromosome is assigned to a cluster (K) based on its rank.

**2. Pre-processing.** This component consists of the following steps:

   A. During the first iteration, the following steps are conducted:

     i. Data cleaning: The data may include many unnecessary and missing elements. Data cleansing is performed to address this, and includes the management of missing and noisy data;

     ii. Data normalisation: This step is used to transform the categorical dataset into numerical data to be suitable for the Euclidean distance clustering process. Normalising data is used to scale the data values. For example, Table 1 presents the categorical data of three attributes.

Table 1. Sample table of categorical data

| x1 | x2 | x3 |
|----|----|----|
| a  | E  | x  |
| b  | D  | x  |
| c  | E  | x  |
| d  | D  | x  |
| e  | E  | x  |
| f  | D  | x  |

After using data normalisation process, the multiple variables presented in Table 1 can be converted to numeric values. Table 2 shows the new data matrix constructed using numeric columns rather than factorial columns.

Table 2. Numerical representation of Table 1

| x1 | x2 | x3 |
|---|---|---|
| 1 | 2 | 1 |
| 2 | 1 | 1 |
| 3 | 2 | 1 |
| 4 | 1 | 1 |
| 5 | 2 | 1 |
| 6 | 1 | 1 |

iii. The elbow method is used to calculate the number of clusters. The variance (sum of squared errors (SSE) inside clusters) is plotted against the number of clusters in the elbow method. The initial few clusters provide a large amount of variation and information, but the information gain decreases with time, and the shape of the graph becomes angular. The optimum number of clusters is determined; this is referred to as the "Elbow criteria". However, this point cannot be permanently established without ambiguity. The elbow technique is utilised in this study as a visual method for determining the consistency of the optimal number of clusters [23,24]. The idea is to determine the number of clusters, add clusters, and then calculate the SSE for each cluster until the maximum number of clusters is determined. Then, by comparing the difference in SSE for each cluster, the most extreme difference in the elbow angle indicates the optimal cluster number. The SSE formula is given by Equation (14):

$$SSE = \sum_{i=1}^{k} \sum_{x_j \in C_i} || x_j - \mu_i ||^2 \tag{14}$$

where $X_j$ is an object in each cluster and $C_i$ is the centroid of the cluster.

The elbow algorithm is presented in Algorithm 3 to calculate the optimal value of clusters.

**Algorithm 3. The Elbow method**

**Input:** K, Dataset
**Output:** The optimal number of clusters (newK)
1 **while** *(true)* **do**
2     K++;
3     Calculate SEE;
4     **if** *(SSE drops dramatically)* **then**
5        $newK := K$;
6        break;
7     end
8 end

B. The number of real clusters is computed using Equation (2);
  C. The initial cluster centroids are calculated using Equation (3);
  D. The old cluster centroids are computed using Equation (4);
  E. For each cluster, the intraCluster and oldIntraCluster are determined. Similarly, for the current clustering solution, the interCluster and oldInterCluster are calculated. Equations (5) and (6) present the mathematical formula of intraCluster and interCluster calculations for clusters A and B, respectively;
  F. Finally, the new cluster centroids are calculated as shown in Equation (7).

**3. Mut-over.** Similar to ECA*, this component consists of mutation and crossover operators. The equations for this step are presented in Section 2.

**4. Post-processing.** Following the generation of the mut-over operator, the following steps are performed:

  A. Merging clusters according to their diversity using Equations (12) and (13);
  B. Recalculating the cluster centroids;
  C. De-normalising the dataset if the termination requirements are met. This step de-normalises the numeric data into the original data (categorical dataset).

**5. Fitness evaluation.** The halting requirements are satisfied if the following conditions are satisfied:

  A. The procedure completes the given number of iterations;
  B. The values of the interCluster and intraCluster remain constant during each cycle. This means that the interCluster value does not increase, and the intraCluster value does not decrease throughout each cycle.

**4. Methods**

The proposed methodology is depicted in Fig. 2, in which COVID-19 and medical disease datasets are used as inputs to the proposed iECA* algorithm and other algorithms. Then, the clustering results are validated using three performance and validation measures. Finally, the success and failure ratios of the methods on each dataset are depicted to demonstrate the performance of each algorithm.

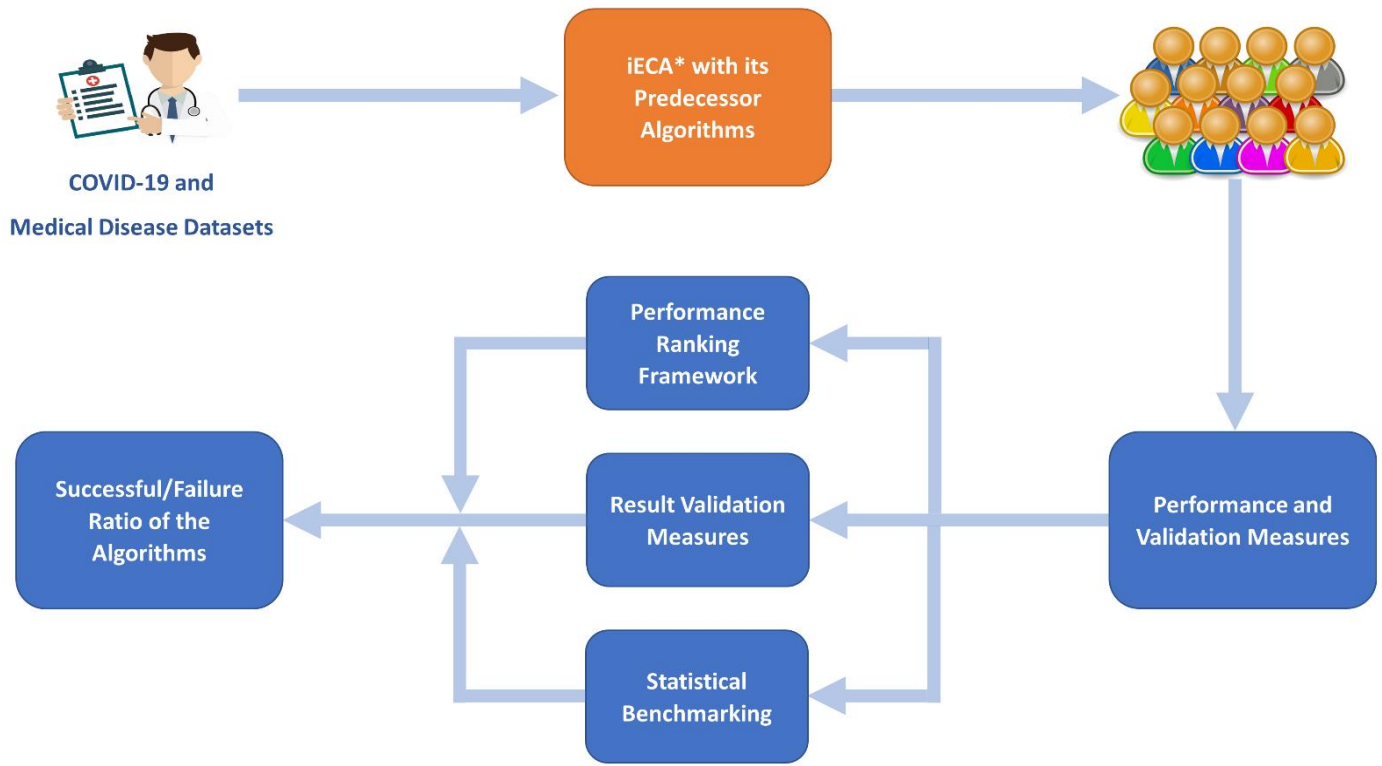

**Fig. 2. The proposed methodology**

### 4.1. Datasets

The real-world datasets of COVID-19 symptoms checker, Liver disorders, Diabetes, Kidney and Heart diseases are used to evaluate the adaptive ECA* against its predecessor algorithms.

**1. The COVID-19 symptoms checker:** This dataset can aid in determining whether an individual has coronavirus disease based on a set of pre-defined typical symptoms. This dataset is taken from [25]. These signs are based on the Indian government and the World Health Organisation (WHO). The findings of these data analyses could be construed as medical recommendations. The dataset includes seven main variables that would affect whether anyone has coronavirus disease. A combination will be created for each mark in the variable with all these categorical variables, resulting in 316800 variations. The whole dataset encompasses raw and cleansed datasets. We have used the cleansed and reprocessed version of the dataset with the same number of instances and attributes.

**2. Medical datasets:** The second sets of data include four real datasets:

A. The Liver disorder's real-world dataset is collected from the UCI Machine Learning Repository and donated by Richard [26]. This dataset consists of 341 instances with seven categorical, integer, and real attributes. The first five factors pertain to blood samples, which are vulnerable to Liver Disorders caused by heavy alcohol intake. Meanwhile, each record in the dataset corresponds to a single male person. Blood samples are used to determine the mean corpuscular volume (MCV), alkaline phosphatase (Alk Phos), alanine aminotransferase (SGPT), aspartate aminotransferase (SGOT), and gamma-glutamyl transpeptidase (Gammage) activity, as well as the number of half-pint equivalents of alcoholic beverages drunk each day (drinks). The 341 samples are divided into two distinct groups based on liver disorders: class 1, which contains 142 samples, and class 2, which contains 199 samples.

B. The Diabetes dataset is collected from the United States of America and Turkey, accessible online [27]. It has 768 instances and 9 data attributes, such as Glucose, Blood pressure, Insulin, Age and Outcome. This dataset aims to check the patients whether have Diabetes or not. This grouping is conducted via the value of the 'Outcome' attribute (1: The patient has diabetes; 0: The patient does not have Diabetes).

C. The Kidney dataset is gathered in India over two months [28]. It contains 400 rows and 25 characteristics, including red blood cells, sugar, and pedal oedema. This data aims to ascertain whether or not a patient has chronic kidney disease. This type is determined by the value of an attribute called 'Group', either chronic kidney disease (CKD) or not-CKD. We have cleansed the dataset, including mapping the text to numbers and making a few other improvements.

D. Additionally, the Heart disease dataset is owned by David Lapp and collected on 04/06/2019 from four different databases (Hungary, Cleveland, Long Beach V, and Switzerland) [29]. It consists of 1025 instances and 14 attributes. This dataset can be clustered based on the presence of heart disease in the instance of patients. Zero means no heart disease, while one means there exist heart disease in the instance record. Table 3 presents the characteristics of the used datasets.

Table 3. Dataset characteristics

| Datasets | Dataset characteristics | Attribute characteristics | Number of instances | Number of attributes | Missing values? | Date donated |
|---|---|---|---|---|---|---|
| COVID-19 symptoms checker | Multivariate, real | Real, Integer, Categorical | 316800 | 7 | N/A | 21/03/2020 |
| Liver disorders | Multivariate, real | Real, Integer, Categorical | 345 | 7 | No | 15/05/1990 |
| Diabetes dataset | Multivariate, real | Real, Integer, Categorical | 769 | 8 | No | 13/06/2020 |
| Kidney disease | Multivariate, real | Real | 400 | 25 | Yes | 03/07/2015 |
| Heart disease | Multivariate, Domain-Theory | Real, Integer Categorical | 1025 | 14 | N/A | 06/06/2019 |

## 4.2. Experimental Setup

We conducted experiments to evaluate the results of iECA* compared to those of seven state-of-the-art methods. The primary objectives of this experiment were: (i) clustering COVID-19 and real-world patient datasets effectively; (ii) evaluating the performance of iECA* using the performance and validation measures of the clustering algorithms.

We run iECA* and its counterpart algorithms 30 times on every dataset to determine the cluster consistency and cluster objective function for each run. The clustering solutions of the algorithms varied between runs. Each run consisted of 50 iterations. For each dataset problem, Weka 3.9 was used to run the counterpart algorithms of iECA*. We also report the average outcomes for each technique for the 30 clustering solutions for each dataset problem. Uniform crossover was used as part of the mut-over strategy, as it is an efficient and powerful operator form in evolutionary algorithms to reduce joint problems [30]. It is also challenging to initialise the optimal value for the pre-defined ECA* variables. iECA* also lacks the correct cluster density threshold to be chosen, which should be sufficient for compound and multi-featured issues. The cluster density threshold may differ depending on the type of benchmarking issue. For example, a cluster density threshold of 0.001 may be suitable for one type of dataset but not for another one. Nonetheless, the cluster density threshold can be calculated based on the scale and characteristics of the dataset. As a result, we conclude that the initial values mentioned in Table 4 are optimal for addressing the dataset issues found during this study. Table 4 presents the criteria and parameters for running iECA*.

**Table 4. The criteria and parameters for running iECA***

| Criteria | Initial value |
| --- | --- |
| Initial number of clusters | 1 |
| Cluster density ratio | 0.001 |
| Alpha (random walk) | 0.001 |
| Maximum number of runs | 30 |
| Maximum iterations | 50 |
| Crossover operator type | Uniform crossover |

As described in Weka 3.9, the primary criteria for running the competitive algorithms of iECA* are presented in Table 5.

Table 5. The primary criteria for running counterpart algorithms of iECA*

| Criteria | Initial value |
|---|---|
| Number of clusters | Depends on the dataset |
| Maximum number of runs | 30 |
| Maximum iterations | 50 |
| Input centre file and debug vectors file | Weka 3.9 |

Additionally, Table 6 provides the parameters used for running each counterpart algorithm of iECA* and the reason for choosing these parameters.

Table 6. The parameters used for running each counterpart algorithm of iECA*

| Algorithms | Parameters | Reasons |
|---|---|---|
| ECA* | Cluster density threshold: 0.001<br>Alpha (random walk): 1.001<br>Number of social class rank: 2-10<br>Type of crossover operator: Uniform crossover | The pre-defined parameters are initialised to be implemented following the dataset's size and characteristics [14]. |
| GENCLUST++ | Number of clusters: climbing hill<br>Initial population size: 30<br>Seed: 10 | We adhere to the initial values suggested by the original publications [31]. |
| LVQ | Number of clusters: Depends on the dataset<br>Learning rate: 1.0<br>Normalise attributes: True | On several problems, the initial parameter settings for the 11 LVQ classifiers showed LVQ's superior performance [32]. |
| SVM | Number of hyperplanes: 2-10<br>Gamma: From 0.0001 to 10<br>C parameter: From 0.1 to 100<br>Batch size: 10 | Many SVM parameters, including the c and gamma parameters, should be selected [33]. The optimum values for these parameters utilised in [34] are employed in this study to reduce the training error. |
| ANN | Input layer size: Depends on the features of the datasets.<br>Hidden layer size: 2<br>Output layer size: 2-4<br>Threshold range: [-1. 1]<br>Weight range: [-1, 1]<br>Learning coefficient: 0.2<br>Activation function: Sigmoid<br>Momentum: 0.8 | These parameters are initialised based on the previous protocol presented in [35]. |
| KNN | Number of neighbourhoods: 3-10<br>Distance function: Hamming distance | Selecting the optimum value for K is best accomplished by examining the data first. A high K value generally results in greater precision since it lowers total noise, although this is not a guarantee. Cross-validation is another technique for determining a suitable K value retroactively by comparing |

| | | it to an independent dataset. Historically, the optimum K value for most datasets was between 3 and 10 [36]. This gives much more accurate results than 1NN. Additionally, it should be emphasised that Euclidean, Manhattan, and Minkowski distance measures are valid only for continuous variables. When categorical variables are included, the hamming distance should be employed [37]. |
|---|---|---|
| Deep KNN | The same parameters of KNN with feature extraction. | Due to the nonparametric nature of KNN, it is challenging to include KNN classification into feature extractor learning. [38] presents an end-to-end learning method for integrating KNN classification and feature extraction. We have utilised the same procedure of the mentioned study since experiments showed that the proposed deep KNN outperforms KNN and other strong classifiers. |

### 4.3. Performance and Validation Measures

### 4.3.1. Validation Measures

According to [39], clustering assessment and validation are almost as crucial as clustering itself. Numerous quality measures and objective function measures are available to evaluate clustering performance. In our study, we utilised five cluster validation measures to evaluate iECA*.

**1. Accuracy:** Accuracy is equal to the ratio of the number of correct matching pairs to the total number of matching pairs. A true positive (TP) result places two pairs connected in the same cluster; a true negative (TN) result places two pairs of dissimilar data points in separate clusters. A false positive (FP) result allocates two data points that are distinct to the same cluster. A false negative (FN) result classifies two similar points to distinct clusters [40]. The accuracy was calculated using Equation (15).

$$Accuracy = \frac{TP + TN}{TP + TN + FP + FN} \quad (15)$$

**2. Normalised mutual information (NMI):** NMI is an external metric for determining the quality of clustering. Because this approach is normalised, we may compare the NMI across clusters with varying numbers of clusters. Consider the set of clusters K, class label C, entropy H(.) and mutual information I (C: K). The NMI is then determined using Equation (16):

$$NMI\ (C,K) = \frac{2\ X\ I\ (C:K)}{[H(C) + H(K)]} \quad (16)$$

where $I(C:K) = H(C) - H((C|K))$.

**3. Adjusted Rand index (ARI):** The ARI uses the global hypergeometric distribution as the random model. In other words, the V and U partitions are randomly chosen such that the number of objects in the clusters remains constant. Let $n_{ij}$ be the total number of items in classes $u_i$ and $v_j$. Consider the numbers $n_i$ and $n_j$ to represent the number of items in classes $u_i$ and $v_j$, respectively [40]. Therefore, ARI is defined in Equation (17):

$$ARI = \frac{\sum_{i,j}(2^{n_{i,j}}) - [\sum_i(n_{i_2}) \sum_j(2^{n_j})]/(2^n)}{\frac{1}{2}[\sum_i(2^{n_i})] - [\sum_i(2^{n_i}) \sum_j(2^{n_j})]/(2^n)} \tag{17}$$

**4. Normalised mean squared error (nMSE):** The mean squared error (MSE) is used to calculate the average of squared errors as well as the average squared difference between the actual and estimated values. As in [2], nMSE was employed in this study, as shown in Equation (18):

$$nMSE = \frac{SSE}{N.D} \tag{18}$$

where *SSE* is the sum of squared errors, *N* is the number of populations, and *D* is the number of attributes in the dataset.

The SSE calculates the squared differences between each observation, its cluster centroid, and the variance within a cluster. If all the cases in a cluster are identical, the SSE is equal to zero. That is, the lower the SSE value, the better the work of the algorithms. For instance, if one method returns an SSE of 7.44, and another returns an SSE of 17.26, we may infer that the former approach performs better than the latter. Equation (14) illustrates the SSE.

**5. Davies–Bouldin index (DBI):** The DBI is a clustering method evaluation measure used to measure the average similarity of each cluster with its most similar cluster. This is an internal assessment method in which the quality of the clustering is determined using dataset-specific variables and characteristics [41]. The MATLAB implementation of DBI is available via the MATLAB Statistics and Machine Learning Toolbox, using the "evalclusters" command [42].

### 4.3.2. Statistical Benchmarking

As part of the performance validation of iECA* against the state-of-the-art algorithms, we evaluated the overall performance of the algorithms in terms of running time against memory usage for each medical dataset. We also compared the average execution time with memory consumption for the 30 solutions obtained by iECA* and the other algorithms.

### 4.3.3. Performance Ranking Framework

To determine the performance ranking level of each algorithm according to each dataset and each performance validation metric (accuracy, NMI, ARI, nMSE, and DBI), we evaluated the effectiveness of iECA* with the current methods in two manners: (i) we assessed the performance of each algorithm with respect to each dataset using all the validation metrics. The ranking level varied from 1 (the best algorithm) to 8 (the worst algorithm). (ii) We ranked the performance of each algorithm for each performance validation metric using all datasets. The ranking level is represented by three colours: green (good performance), yellow (moderate performance), and red (poor performance).

## 5. Results

This section is divided into three sub-sections: performance result analysis, statistical performance benchmarking, and performance rating framework.

### 5.1. Performance Result Analysis

The accuracy of all five datasets is shown in Fig. 3. The results obtained indicate that the suggested iECA* method outperformed current algorithms such as ECA* in terms of the validation controls utilised (accuracy, NMI, ARI, nMSE, and DBI) in nearly all situations. The iECA* algorithm improved the accuracy by 3.5% in the COVID-19 symptoms checker and kidney datasets and by 4.7% in the liver disorder dataset. On the diabetes dataset, the accuracy was almost the same as that of the current methods, whereas it was improved by 4.5% in the heart disease dataset.

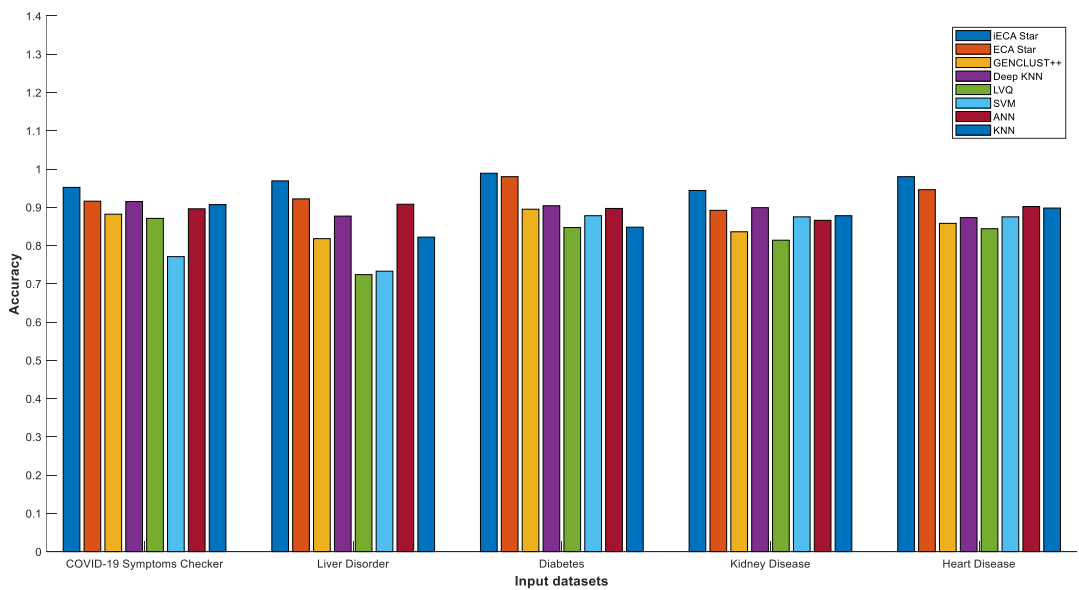

**Fig. 3. Accuracy results of iECA* compared to other algorithms on COVID-19 and medical disease datasets**

The NMI comparison is shown in Fig. 4. As a result, we conclude that the proposed clustering method provides superior performance in all cases. The NMI value of iECA* fully agreed with the ground truth results for the current liver disorder, diabetes, and heart disease datasets. In addition, there was a relative increase of 1% in the NMI for COVID-19 symptom checker and kidney disease datasets with a slight difference.

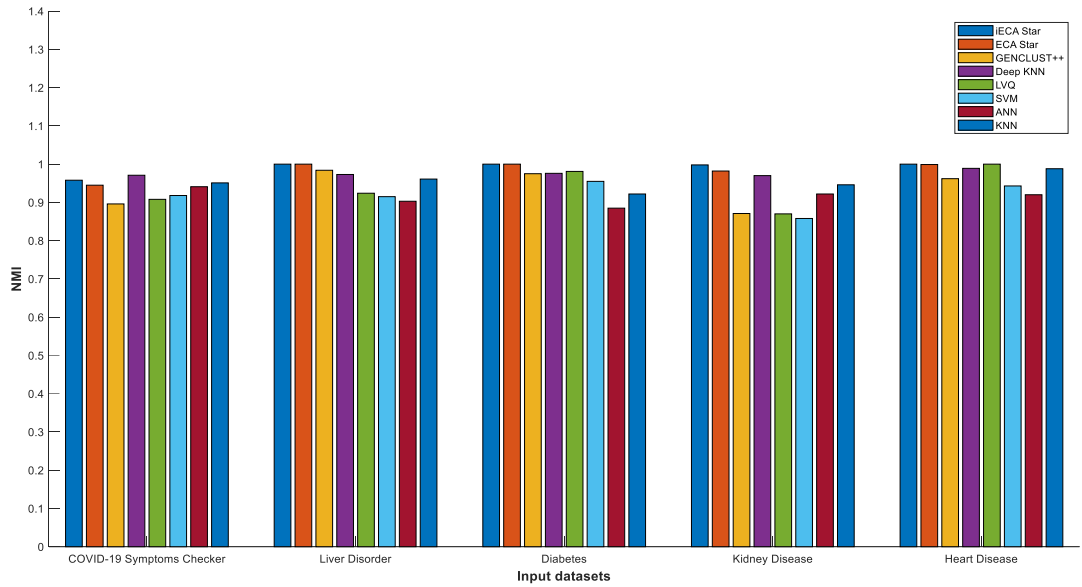

**Fig. 4. NMI results of iECA* compared to other algorithms on COVID-19 and medical disease datasets**

Fig. 5 shows the ARI comparison. The iECA* method outperformed the previous data clustering algorithms, providing an increase of 1% in the ARI values in the COVID-19 symptoms checker, liver disorder, and kidney disease datasets. The suggested method provided similar results as those of the current techniques in terms of ARI for diabetes and heart disease datasets.

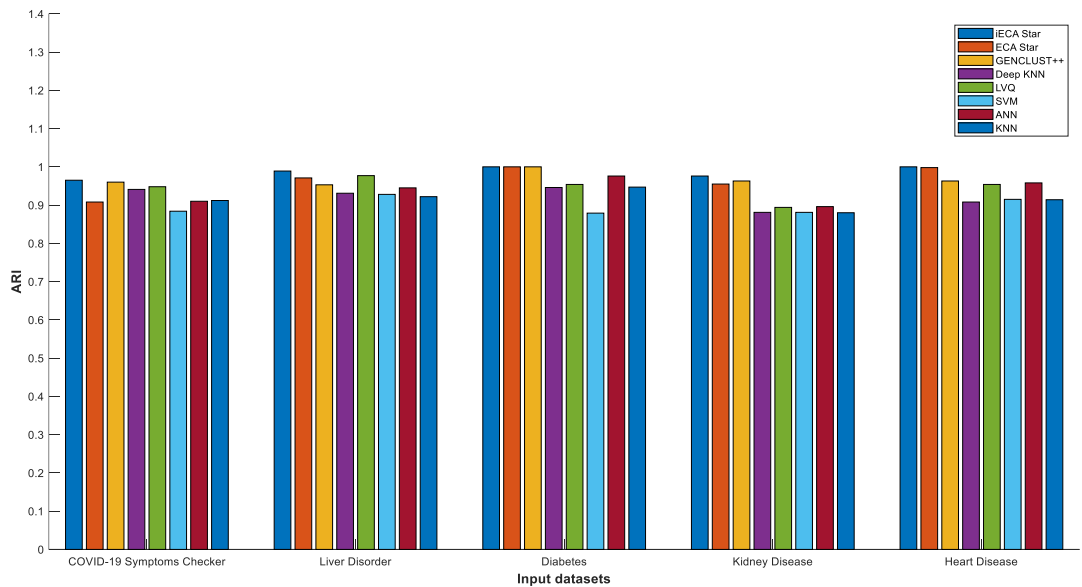

**Fig. 5. ARI results of iECA* compared to other algorithms on COVID-19 and medical disease datasets**

In terms of nMSE, in all datasets except the diabetes dataset, iECA* outperformed the other approaches analysed. In contrast, in the diabetes dataset, ANN had superior performance, followed by deep KNN. Fig. 6 shows the nMSE results.

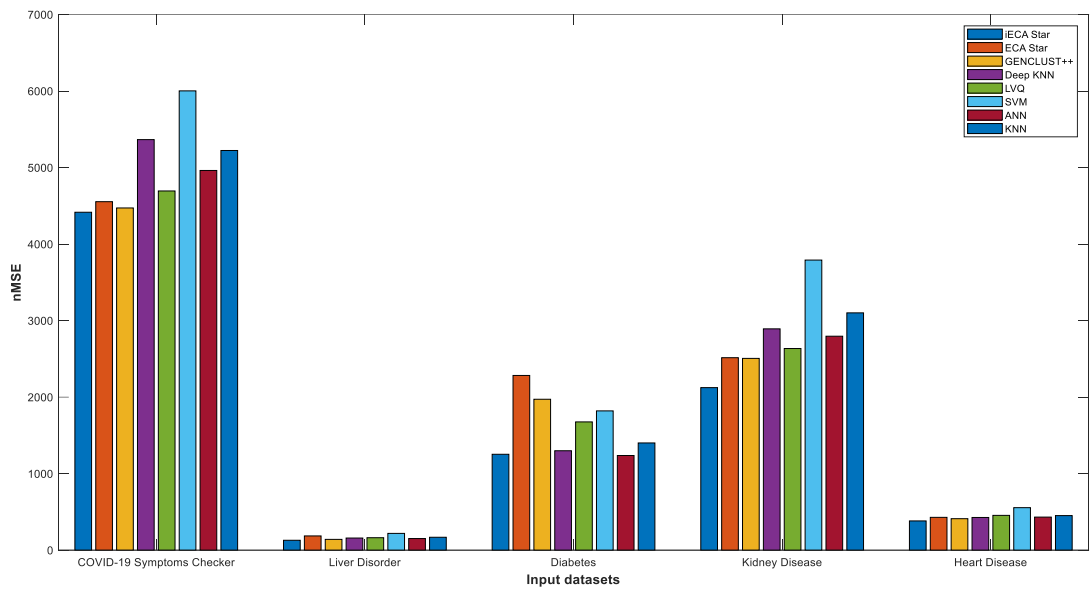

**Fig. 6. nMSE results of iECA* compared to other algorithms on COVID-19 and medical disease datasets**

Fig. 7 presents the DBI comparison. It is observed that the iECA* outperformed the current data clustering methods, providing an improvement of 3% for the COVID-19 symptoms checker and liver disorder datasets, 2% for diabetes and kidney disease datasets, and 5% for the heart disease dataset compared to the other data clustering approaches.

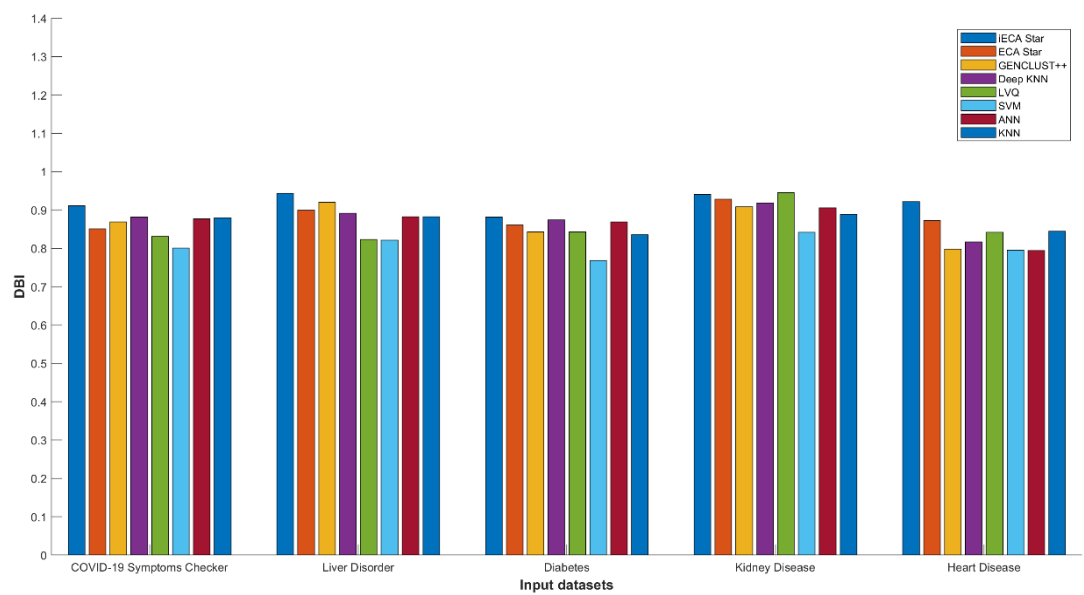

**Fig. 7. DBI results of iECA* compared to other algorithms on COVID-19 and medical disease datasets**

## 5.2. Statistical Performance Benchmarking

This section analyses the overall performance benchmarking of the algorithms (execution time/memory consumption) based on the datasets. Table 7 presents the execution time with memory consumption for the 30 solutions obtained by iECA* and other algorithms. We observe that iECA* exhibited a shorter execution time for clustering all the datasets. Similarly, the proposed method consumed less memory than the other techniques. Surprisingly, on the kidney disease dataset, iECA* required a higher memory allocation than deep KNN. In general, the proposed iECA* technique had a faster execution and consumed less memory than the other clustering methods.

Table 7. Average execution time with memory consumption for the 30 solutions obtained by iECA* and other algorithms

| Datasets | Benchmarking criteria | iECA* | ECA* | GENCLUST++ | Deep KNN | LVQ | SVM | ANN | KNN | Winner |
|---|---|---|---|---|---|---|---|---|---|---|
| COVID-19 symptoms checker | Execution time | **567.129** | 601.484 | 724.937 | 589.937 | 701.937 | 665.937 | 711.847 | 602.873 | iECA* |
| | Memory consumption | **86.134** | 96.283 | 142.928 | 104.918 | 151.928 | 98.283 | 102.817 | 114.273 | iECA* |
| Liver disorder | Execution time | **48.273** | 54.1832 | 50.384 | 50.283 | 64.184 | 61.827 | 49.173 | 55.183 | iECA* |
| | Memory consumption | **19.481** | 23.168 | 31.384 | 26.184 | 29.474 | 26.833 | 29.857 | 28.437 | iECA* |
| Diabetes | Execution time | **71.273** | 82.383 | 89.638 | 79.1737 | 96.174 | 94.8173 | 88.371 | 83.239 | iECA* |
| | Memory consumption | **36.244** | 39.172 | 51.761 | 44.183 | 50.819 | 49.128 | 41.347 | 47.127 | iECA* |
| Kidney disease | Execution time | **38.284** | 42.8173 | 54.183 | 46.283 | 53.173 | 51.827 | 49.718 | 43.718 | iECA* |
| | Memory consumption | 18.177 | 21.661 | 29.384 | **17.987** | 23.817 | 24.981 | 19.384 | 18.341 | Deep KNN |
| Heart disease | Execution time | **79.274** | 84.134 | 93.193 | 87.283 | 94.184 | 97.287 | 91.827 | 90.073 | iECA* |
| | Memory consumption | **42.387** | 46.126 | 56.128 | 48.128 | 54.128 | 53.383 | 45.651 | 50.841 | iECA* |

Additionally, Fig. 8 illustrates the average execution time for the 30 solutions obtained by iECA* and its competitive algorithms.

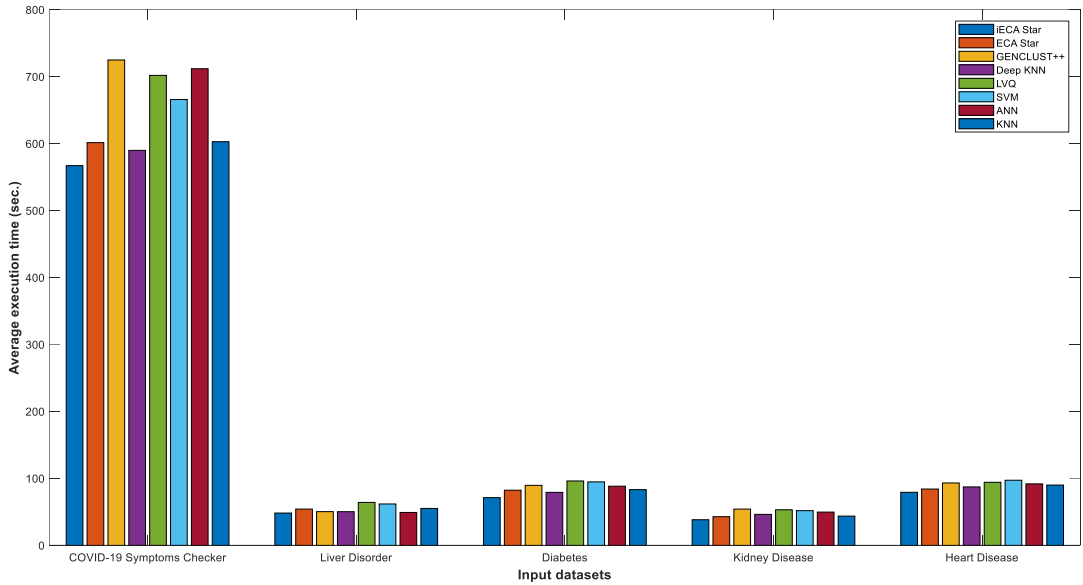

**Fig. 8. Average execution time for the 30 solutions obtained by iECA* and other algorithms**

Furthermore, Fig. 9 illustrates the average memory consumption for the 30 solutions obtained by iECA* and its competitive algorithms.

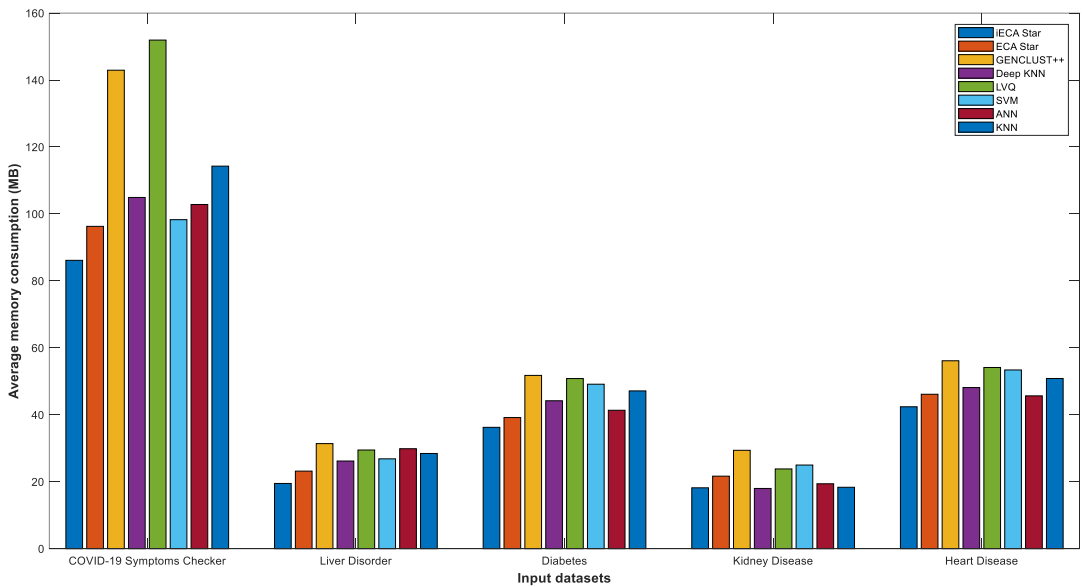

**Fig 9. Average memory consumption for the 30 solutions obtained by iECA* and other algorithms**

## 5.3. Performance Ranking Framework

We ranked the algorithms according to their effectiveness on the five datasets according to the clustering validation measure. The ranking level ranged from 1 (the best algorithm) to 8 (the worst algorithm).

Table 8 presents the ranking of the algorithms for the COVID-19 symptom checker. We notice that the iECA* scored 1.2 on average, followed by deep KNN, KNN and ECA*. Conversely, SVM was the worst algorithm for clustering the COVID-19 dataset.

Table 8. Ranking level of the algorithms for the COVID-19 symptoms checker dataset

| Criteria/ Statistics | iECA* | ECA* | GENCLUST++ | Deep KNN | LVQ | SVM | ANN | KNN |
|---|---|---|---|---|---|---|---|---|
| Accuracy | 1 | 2 | 6 | 3 | 7 | 8 | 5 | 4 |
| NMI | 2 | 4 | 8 | 1 | 7 | 6 | 5 | 3 |
| ARI | 1 | 6 | 2 | 4 | 3 | 7 | 6 | 5 |
| nMSE | 1 | 3 | 2 | 7 | 3 | 8 | 5 | 6 |
| DBI | 1 | 6 | 5 | 2 | 7 | 8 | 4 | 3 |
| Average | 1.20 | 4.20 | 4.60 | 3.40 | 5.40 | 7.40 | 5.00 | 4.20 |

For the liver disease dataset, iECA* outperformed all other algorithms, whereas SVM failed to surpass all the others. Both ECA* and GENCLUST++ had an average rank of 3.4. The rank of the algorithm and total rank are listed in Table 9.

Table 9. Ranking level of the algorithms for the liver disorder dataset

| Criteria/ Statistics | iECA* | ECA* | GENCLUST++ | Deep KNN | LVQ | SVM | ANN | KNN |
|---|---|---|---|---|---|---|---|---|
| Accuracy | 1 | 2 | 6 | 4 | 8 | 7 | 3 | 5 |
| NMI | 1 | 2 | 3 | 4 | 6 | 7 | 8 | 5 |
| ARI | 1 | 3 | 4 | 6 | 2 | 7 | 5 | 8 |
| nMSE | 1 | 7 | 2 | 4 | 5 | 8 | 3 | 6 |
| DBI | 1 | 3 | 2 | 4 | 7 | 8 | 6 | 5 |
| Average | 1.00 | 3.40 | 3.40 | 4.40 | 5.60 | 7.40 | 5.00 | 5.80 |

iECA* had an average rank of 1.2 in the diabetes dataset, followed by ECA* and deep KNN. Table 10 summarises the criteria and the ranking of each method for the diabetes dataset.

Table 10. Ranking level of the algorithms for the diabetes dataset

| Criteria/ Statistics | iECA* | ECA* | GENCLUST++ | Deep KNN | LVQ | SVM | ANN | KNN |
|---|---|---|---|---|---|---|---|---|
| Accuracy | 1 | 2 | 5 | 3 | 8 | 6 | 4 | 7 |
| NMI | 1 | 2 | 5 | 4 | 3 | 6 | 8 | 7 |
| ARI | 1 | 2 | 3 | 7 | 5 | 8 | 4 | 6 |
| nMSE | 2 | 8 | 7 | 3 | 5 | 6 | 1 | 4 |
| DBI | 1 | 4 | 5 | 2 | 6 | 8 | 3 | 7 |
| Average | 1.20 | 3.60 | 5.00 | 3.80 | 5.40 | 6.80 | 4.00 | 6.20 |

Additionally, Table 11 provides the ranking level of the algorithms for the kidney disease dataset. On average, iECA* was a superior clustering method for kidney data, followed by ECA*, deep KNN, GENCLUST++, LVQ, ANN, KNN, and SVM.

Table 11. Ranking level of the algorithms for the kidney disease dataset

| Criteria/ Statistics | iECA* | ECA* | GENCLUST++ | Deep KNN | LVQ | SVM | ANN | KNN |
|---|---|---|---|---|---|---|---|---|
| Accuracy | 1 | 3 | 7 | 2 | 8 | 5 | 6 | 4 |
| NMI | 1 | 2 | 6 | 3 | 7 | 8 | 5 | 4 |
| ARI | 1 | 3 | 2 | 6 | 5 | 7 | 4 | 8 |
| nMSE | 1 | 3 | 2 | 6 | 4 | 8 | 5 | 7 |
| DBI | 2 | 3 | 5 | 4 | 1 | 8 | 6 | 7 |
| Average | 1.20 | 2.80 | 4.40 | 4.20 | 5.00 | 7.20 | 5.20 | 6.00 |

The ranking levels for the heart disease dataset are listed in Table 12. On average, iECA* was the best algorithm, followed by ECA*. GNELCUST++, deep KNN, KNN, and LVQ performed similarly well. Nonetheless, ANN and SVM algorithms were the least effective.

Table 12. Ranking level of the algorithms for the heart disease dataset

| Criteria/ Statistics | iECA* | ECA* | GENCLUST++ | Deep KNN | LVQ | SVM | ANN | KNN |
|---|---|---|---|---|---|---|---|---|
| Accuracy | 1 | 2 | 7 | 6 | 8 | 5 | 3 | 4 |
| NMI | 1 | 3 | 6 | 4 | 2 | 7 | 8 | 5 |
| ARI | 1 | 2 | 3 | 8 | 5 | 6 | 4 | 7 |
| nMSE | 1 | 4 | 2 | 3 | 7 | 8 | 5 | 6 |
| DBI | 1 | 2 | 6 | 5 | 4 | 7 | 8 | 3 |
| Average | 1.00 | 2.60 | 4.80 | 5.20 | 5.20 | 6.60 | 5.60 | 5.00 |

Generally, we empirically assessed the performance of these algorithms in a framework over the five datasets according to the five cluster validation measures (accuracy, NMI, ARI, nMSE, and DBI). Fig. 10 depicts the outcome rating scale for iECA* compared to five real-world patient datasets. The values presented in Fig. 10 are aggregated from the average ranking level of the algorithms presented in Tables 8-12 for the COVID-19 symptoms checker, liver disorder, diabetes, kidney disease, and heart disease datasets. The green colours indicate that the algorithm performed well (ranked first) for a particular dataset value. The red colours indicate that the technique exhibited poor performance (ranked as a third class). The yellow colours represent that the current technique performed moderately for its corresponding medical data (ranked as a second class). Specifically, the colour areas were numbered from 1 (green) to 8 (red) inclusively as follows:

- Green: from 1.000 to 3.332 (good performance).
- Yellow: from 3.333 to 5.665 (moderate performance).
- Red: from 5.666 to 8.000 (poor performance).

We interpret those values with diverging scales of colour to demonstrate colour development in two directions [43]: progressively toning down the first hue from one end to a neutral colour at the midway, then increasing the opacity of the second hue to the other end.

| Criteria/ Statistics | iECA star | ECA star | GENCLUST++ | Deep KNN | LVQ | SVM | ANN | KNN |
|---|---|---|---|---|---|---|---|---|
| COVID-19 symptoms checker | 1.20 | 4.20 | 4.60 | 3.40 | 5.40 | 7.40 | 5 | 4.2 |
| Liver disorder | 1.00 | 3.40 | 3.40 | 4.40 | 5.60 | 7.40 | 5 | 5.8 |
| Diabetes | 1.20 | 3.60 | 5.00 | 3.80 | 5.40 | 6.80 | 4 | 6.2 |
| Kidney disease | 1.20 | 2.80 | 4.40 | 4.20 | 5.00 | 7.20 | 5.2 | 6 |
| Heart disease | 1.00 | 2.60 | 4.80 | 5.20 | 5.20 | 6.60 | 5.6 | 5 |
| Average | 1.12 | 3.32 | 4.44 | 4.20 | 5.32 | 7.08 | 4.96 | 5.44 |

Fig. 10. Heatmap for the performance ranking framework of iECA* (Green: good performance; Yellow: moderate performance; Red: poor performance)

The findings indicate that iECA* outperformed the other algorithms in clustering all medical datasets, followed by ECA* and deep KNN. GENCLUST++ was the fourth most successful algorithm, with an average score of 4.44. Deep KNN, ANN, and GENCLUST++ were considered algorithms with reasonable performance to cluster all the datasets, whereas SVM and KNN could not detect most of the clusters of the medical data. As stated, iECA* did not outperform the other algorithms in a few cases.

There are two main reasons for this result. First, according to the no-free-lunch theorems [44], any algorithm that performs exceptionally well on one set of objective functions (datasets) must perform poorly on all other sets. Other factors, such as the cohort of the problem, type of dataset, and difficulty of the problem, might affect the performance of an algorithm on a specific type of problem [6]. This means that a definitive evaluation about the absolute success of iECA* and other algorithms in grouping dataset issues cannot be made solely on their difficulty scores. As a result, there is no inherent connection between the performance of these algorithms and the complexity of clustering medical datasets. Overall, for all five fundamental data properties, the algorithms were ranked as follows: iECA*, ECA*, deep KNN, GENCLUST++, ANN, LVQ, KNN, and SVM.

## 6. Conclusions

In this study, we proposed iECA* by (i) utilising the elbow method to determine optimal cluster numbers and (ii) cleaning and processing data as part of the algorithm. iECA* was utilised to cluster real datasets of COVID-19 and other medical diseases. We also evaluated iECA* based on the aforementioned datasets and compared it with seven other modern clustering algorithms. The evaluation process was conducted for iECA* using five cluster validation measures (accuracy, NMI, ARI, nMSE, and DBI), statistical benchmarking in running time against memory usage, and performance ranking. Three significant findings emerged from the evidence of experimental studies. First, iECA* outperformed the other competing algorithms in clustering the selected medical disease datasets using cluster validation criteria. Second, iECA* outperformed the existing clustering algorithms in terms of execution time and memory usage for clustering all datasets. Third, an operational methodology was proposed to compare the efficacy of iECA* with that of other algorithms in the datasets analysed. The framework showed that iECA* exhibited a better performance compared to the other algorithms in all medical datasets. ECA* was ranked as the second-best algorithm, followed by deep KNN. Following these three successful algorithms, GENCLUST++ was ranked fourth. Deep KNN, ANN, and GENCLUST++ were considered as methods with a reasonable performance for clustering all datasets, whereas SVM and KNN were unable to identify the majority of clusters in the five medical datasets. Thus, the methods were ranked as follows for the five essential datasets: iECA*, ECA*, deep KNN, GENCLUST++, ANN, LVQ, KNN, and SVM.

The main values of iECA* over its counterpart algorithms are five-fold: (i) the elbow technique is used to determine the optimal number of clusters. Perhaps the most well-known technique for finding the optimal cluster number is the elbow method. This is a heuristic technique for estimating the number of clusters in each dataset in cluster analysis. (ii) the input dataset is cleaned and pre-processed to remove unnecessary and missing elements and transform the categorical dataset into numerical data suitable for the Euclidean distance clustering process. (iii) the output dataset is post-processed to de-normalise the numeric data into the original data (categorical dataset). (iv) unlike ECA*, iECA* applies to multivariate and domain-theory real datasets with various attribute characteristics, including integer, real, and categorical data attributes; (v) iECA* was used in real-world clustering applications.

For further research in the future, iECA* can be used for experimental verification of real-world multi-dimensional datasets containing complex knowledge fields to explore more deeply the advantages and drawbacks of the algorithm or improve its efficiency. In addition, iECA* can be applied to more complex and real-world applications to further validate its efficiency, such as engineering application problems [45], library management [46], e-organisation services [47], online analytical processing [48], web engineering [49], and ontology learning [50].

**CRediT author statement**

**Bryar A. Hassan:** Conceptualisation, Methodology, Software, Writing - Original Draft, Visualisation, Formal Analysis, Validation, Methodology. **Tarik A. Rashid:** Project administration, Investigation, Data Curation Writing - Review & Editing, Supervision, Data Curation Writing - Review & Editing. **Hozan Khalid Hamarashid:** Resources, Data Curation Writing - Review & Editing, Writing - Review & Editing, Funding acquisition (if applicable).

**Acknowledgements**

The authors would like to express their heartfelt appreciation to Kurdistan Institution for Strategic Studies and Scientific Research, University of Kurdistan-Hewler, and Sulaimani Polytechnic University for providing facilities and ongoing support for conducting this research.

**Compliance with Ethical Standards**

**Conflict of Interest:** The writers state that they are not involved in any conflict of interest.

**Funding:** The details on funding is inapplicable / No funding was obtained.